\documentclass[pdflatex,sn-nature]{sn-jnl}
\usepackage{graphicx}%
\usepackage{multirow}%
\usepackage{amsmath,amssymb,amsfonts}%
\usepackage{amsthm}%
\usepackage{mathrsfs}%
\usepackage[title]{appendix}%
\usepackage{xcolor}%
\usepackage{textcomp}%
\usepackage{manyfoot}%
\usepackage{booktabs}%
\usepackage{algorithm}%
\usepackage{algorithmicx}%
\usepackage{algpseudocode}%
\usepackage{listings}%
\usepackage{float}%

\theoremstyle{thmstyleone}%
\theoremstyle{thmstyletwo}%
\usepackage{orcidlink}
\theoremstyle{thmstylethree}%
\begin{document}
\title[Article Title]{Automated stereotactic radiosurgery planning using a human-in-the-loop reasoning large language model agent}

\author*[1,2]{\fnm{Humza} \sur{Nusrat}}\email{hnusrat1@hfhs.org}
\author[1,3]{\fnm{Luke} \sur{Francisco}}
\author[1]{\fnm{Bing} \sur{Luo}}
\author[1,2]{\fnm{Hassan} \sur{Bagher-Ebadian}}
\author[1]{\fnm{Joshua} \sur{Kim}}
\author[1]{\fnm{Karen} \sur{Chin-Snyder}}
\author[1,2]{\fnm{Salim} \sur{Siddiqui}}
\author[1,2]{\fnm{Mira} \sur{Shah}}
\author[1,2]{\fnm{Eric} \sur{Mellon}}
\author[4]{\fnm{Mohammad} \sur{Ghassemi}}
\author[1]{\fnm{Anthony} \sur{Doemer}}
\author[1,2]{\fnm{Benjamin} \sur{Movsas}}
\author[1,2]{\fnm{Kundan} \sur{Thind}}

\affil*[1]{\orgdiv{Department of Radiation Oncology}, \orgname{Henry Ford Health}, \orgaddress{\city{Detroit}, \state{MI}}} \affil[2]{\orgdiv{Department of Radiology}, \orgname{Michigan State University}, \orgaddress{\city{East Lansing}, \state{MI}}} \affil[3]{\orgdiv{Department of Statistics}, \orgname{University of Michigan}, \orgaddress{\city{Ann Arbor}, \state{MI}}} \affil[4]{\orgdiv{College of Engineering}, \orgname{Michigan State University}, \orgaddress{\city{East Lansing}, \state{MI}}}

\abstract{Stereotactic radiosurgery (SRS) demands precise dose shaping around critical structures, yet black-box AI systems have limited clinical adoption due to opacity concerns. We tested whether chain-of-thought reasoning improves agentic planning in a retrospective cohort of 41 patients with brain metastases treated with 18 Gy single-fraction SRS. We developed SAGE (Secure Agent for Generative Dose Expertise), an LLM-based planning agent for automated SRS treatment planning. Two variants generated plans for each case: one using a non-reasoning model, one using a reasoning model. The reasoning variant showed comparable plan dosimetry relative to human planners on primary endpoints (PTV coverage, maximum dose, conformity index, gradient index; all $p$ \textgreater\ 0.21) while reducing cochlear dose below human baselines ($p$ = 0.022). When prompted to improve conformity, the reasoning model demonstrated systematic planning behaviors including prospective constraint verification (457 instances) and trade-off deliberation (609 instances), while the standard model exhibited none of these deliberative processes (0 and 7 instances, respectively). Content analysis revealed that constraint verification and causal explanation concentrated in the reasoning agent. The optimization traces serve as auditable logs, offering a path toward transparent automated planning.}

\maketitle
\setlength{\parskip}{0.5em}
\section{Introduction}\label{sec1}
Treatment planning in radiation oncology has grown increasingly complex. A treatment plan consists of a set of instructions generated within a treatment planning system (TPS) that directs the linear accelerator during dose delivery. This task, performed by dosimetrists and medical physicists, requires specialized expertise, substantial time investment, and is subject to variability based on individual planner skill. The degree of complexity varies considerably by tumor site and treatment technique. For conventionally fractionated treatments of anatomically stable sites such as prostate cancer, target volumes are relatively homogeneous and organs at risk occupy predictable positions, facilitating more standardized planning approaches \cite{r1,r2}.

In contrast to this, stereotactic radiosurgery (SRS) for brain metastases represents the opposite end of this spectrum. SRS delivers a large radiation dose to intracranial tumors in a single treatment fraction \cite{r3}. The targets are typically brain metastases, which often present with clinical urgency and compressed treatment timelines. Several factors contribute to the technical difficulty of SRS planning: critical OARs in close proximity to targets, the need for steep dose gradients to minimize normal brain exposure, and the extra precision required when the entire prescribed dose is delivered in one session. Given the current shortage of qualified treatment planners \cite{r4,r5} and the specialized nature of SRS, these treatments are largely confined to large academic medical centers \cite{r6,r7}. Automated treatment planning using artificial intelligence (AI) offers a potential solution to improve access and reduce workforce burden. Prior work in AI-driven treatment planning has largely relied on neural networks trained on institutional retrospective data for specific tumor sites. Several groups have reported successful implementations of this approach \cite{r8,r9,r10}. However, these methods have notable limitations. They are constrained to the anatomical site and treatment technique represented in the training data. Such systems function as black boxes that offer limited transparency or explainability regarding their optimization decisions \cite{r11,r12,r13,r14,r15,r16,r17,r18,r19,r20,r21,r22,r23}. This approach does not scale well across institutions because each implementation remains siloed to the center that performed the development and training. 

Regulatory frameworks for AI-based medical devices increasingly emphasize interpretability and transparency \cite{r24,r25,r26,r27,r28}, while surveys of radiation oncology professionals identify explainability as a key determinant of clinical acceptance \cite{r29,r30,r31}. These concerns around model opacity represent substantial barriers to widespread adoption.

The application of large language models (LLMs) to radiation oncology is an emerging area of investigation. Published work has primarily focused on retrieval-augmented generation (RAG) for clinical question answering, protocol compliance verification, and knowledge-grounded decision support \cite{r32,r33,r34}. These applications leverage LLMs' ability to retrieve and synthesize information from clinical guidelines.

By contrast, the present work employs LLMs for iterative, reasoning-driven treatment plan optimization, a fundamentally different task requiring spatial reasoning, constraint satisfaction, and forward simulation of dosimetric consequences. This distinction is critical: retrieval-based systems improve reliability by grounding responses in validated knowledge sources, whereas reasoning-based planning requires the model to perform multi-step logical inference over complex geometric and dosimetric trade-offs. To date, all reported LLM-based planning studies have employed non-reasoning models without explicit reasoning capabilities, and none have addressed the geometric complexity of SRS planning where transparent, stepwise reasoning is essential. 

The present work addresses these gaps. We employ SAGE, a general agentic framework previously validated for prostate cancer planning \cite{r35}, and apply it to SRS. We directly compare a reasoning LLM against a non-reasoning LLM within the same planning framework and tasks. We include a mechanistic dialogue analysis that connects model behavior to planning outcomes, providing insight into how reasoning architecture influences optimization strategy.

Recent LLM development has produced models optimized for different computational behaviors. While all LLMs fundamentally operate through next-token prediction, some models are specifically trained to generate extended intermediate reasoning steps during inference before producing final outputs. For practical purposes, we refer to these as "reasoning models" versus "non-reasoning models," recognizing this as a behavioral distinction rather than a fundamental architectural dichotomy \cite{r36,r37,r38}.

This behavioral difference, at a functional level, resembles Kahneman's distinction between System 1 (fast, automatic) and System 2 (slow, deliberative) thinking \cite{r39}. We stress that this is an analogy drawn from observable behavior; it is not a claim of cognitive equivalence between artificial and human intelligence. Dual-process theory has shaped how clinical medicine understands complex decision-making. System 2 thinking, with its explicit hypothesis testing, constraint checking, and iterative refinement, appears essential for diagnostic and therapeutic tasks that resist pattern-based shortcuts \cite{r40,r41,r42,r43,r44,r45}. SRS planning shares several characteristics with tasks that empirically benefit from deliberative processing: tightly coupled geometric constraints, competing objectives, three-dimensional spatial reasoning. These properties set SRS apart from more stereotyped planning workflows.

We hypothesized, given these task characteristics, that SRS planning would particularly benefit from LLM architectures exhibiting deliberative, multi-step reasoning behavior. A second hypothesis followed from the first. Intermediate reasoning traces would serve a dual purpose: improving the model's spatial reasoning through self-prompting mechanisms documented in 3D reasoning tasks and constituting an auditable decision log. This log, a structured record of constraint verification, trade-off evaluation, and iterative refinement, can be reviewed by human planners, incorporated into quality assurance documentation, and examined in the event of adverse outcomes.

\section{Methods}\label{sec2}
\subsection{SAGE architecture}\label{subsec1}
The software architecture of SAGE is shown in Figure 1. Upon initialization, the agent receives the clinical scenario (patient anatomy, target volume location and size, spatial relationship between PTV and organs at risk), prescription dose (18 Gy to the PTV in a single fraction), and current state of the optimizer including all dosimetric parameters (DVH metrics for PTV and all OARs). SAGE is then prompted to achieve target coverage while respecting OAR constraints. We tested two variants: a non-reasoning model and a reasoning model. Throughout this manuscript, we use "non-reasoning model" to refer to general-purpose LLMs that generate responses through direct next-token prediction, in contrast to reasoning models that produce intermediate chain-of-thought steps.

Both variants performed up to ten (SAGE stops once clinical goals are met) iterations of optimization, dose calculation, and plan evaluation. Both variants were subjected to identical, deterministic stopping logic. SAGE terminated optimization when all of the clinical goals were simultaneously satisfied. If these criteria were not met after ten iterations, optimization terminated, and the best-performing plan was selected. 

The human-in-the-loop stage served as the decision point for plan disposition. At this stage, a human reviewer either accepted the plan or redirected it to SAGE for a secondary refinement step focused on improving dose conformity. This two-stage architecture allowed us to evaluate both autonomous planning capability and responsiveness to human feedback.

\subsection{Human-in-the-loop refinement}\label{subsec2}
A single board-certified medical physicist evaluated all plans. Plans were accepted if they met all quantitative clinical criteria; those failing to meet conformity benchmarks were directed to the refinement stage.

Refinement followed a standardized protocol. All plans requiring refinement received an identical natural language prompt requesting conformity improvement while maintaining target coverage and OAR constraints. This prompt was applied uniformly across cases regardless of model variant. No case-specific modifications were permitted.

\begin{figure}[H]
  \centering
  \includegraphics[width=0.9\textwidth]{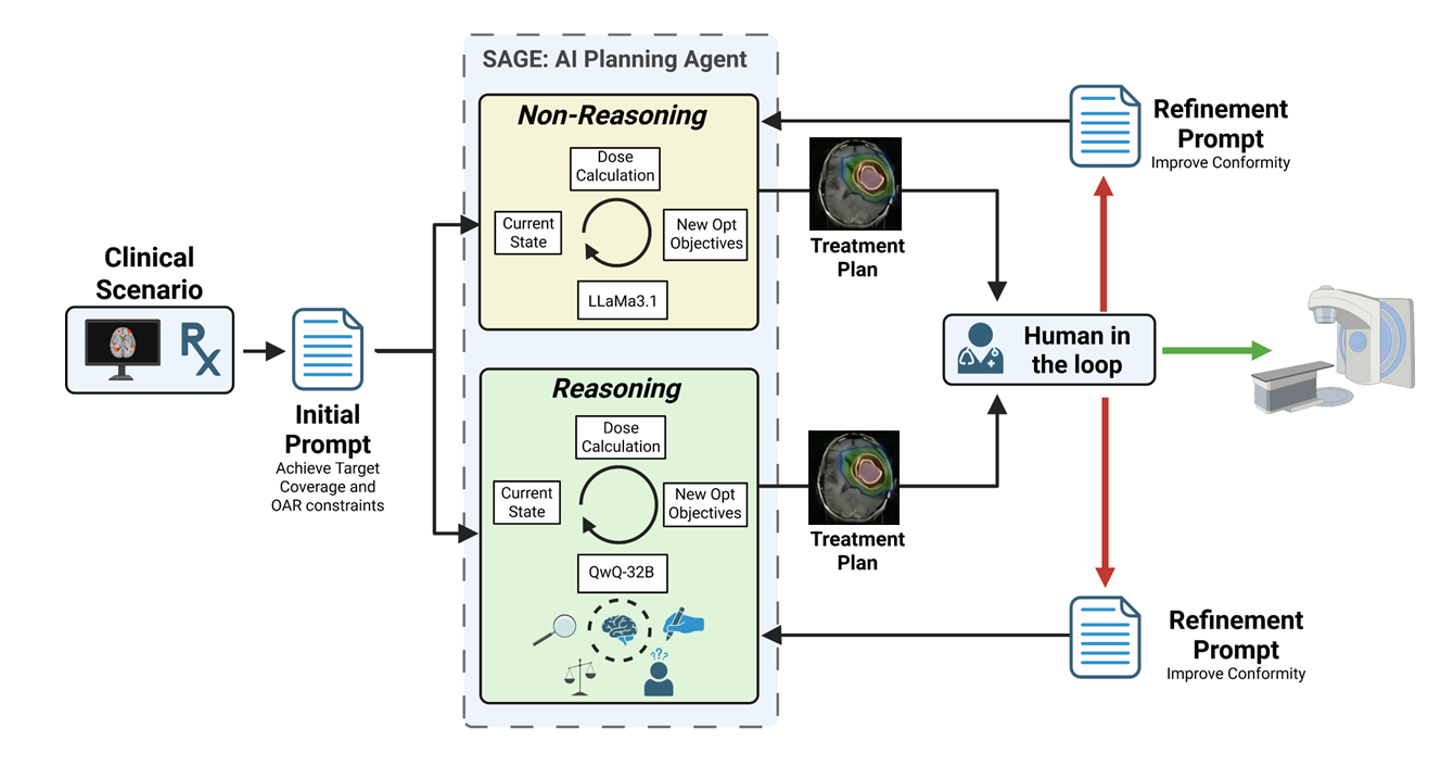}
  \caption{The agent receives clinical inputs (patient anatomy, prescription, physician constraints) and current optimizer state. Two model variants are shown: non-reasoning (top) and reasoning (bottom). Each executes iterative optimization cycles comprising LLM-based parameter adjustment, dose calculation, plan evaluation, and objective updates. The optimal plan proceeds to human review, where it is either accepted or returned with refinement feedback.}
  \label{fig:1}
\end{figure}

\subsection{Model Details}\label{subsec2}
The non-reasoning model was Llama 3.1-70B \cite{r46}. The reasoning model was Qwen QwQ-32B-Reasoning \cite{r47}. LLM hyperparameters were held constant across all experiments (top k = 2; T = 0.4); these values were optimized in prior work by our group \cite{r35}. RAG was enabled, allowing SAGE to access its previous priority number selections and their resulting dose distributions. Both models were hosted locally on eight NVIDIA A100 GPUs within an institutional high-performance computing cluster. 

\subsection{Patient Cohort}\label{subsec3}
We retrospectively obtained treatment data for patients with brain metastases who received single target SRS at our institution between 2022 and 2024. All treatments followed institutional clinical practice guidelines with a prescription of 18 Gy in a single fraction \cite{r48,r49}. The cohort comprised 41 patients for whom CT images, segmented structures, clinical treatment plans, and dosimetric data were available. This study was conducted with institutional review board approval. 

All retrospective clinical plans and SAGE-generated plans were created and housed within the Varian Eclipse Treatment Planning System (version 16.1). Dose calculations were performed using the AAA algorithm (version 15.6.06) with a dose grid resolution of 1.25 mm. For each patient, beam geometry was fixed to match the clinical plan configuration. Dose volume histogram (DVH) estimation (version 15.6.05) and photon optimization (version 15.6.05) algorithms were used for all cases.

\subsection{Mechanistic Content Analysis}\label{subsec4}
To characterize behavioral differences between the reasoning and non-reasoning models beyond dosimetric endpoints, we performed a systematic content analysis of the optimization dialogues generated during planning. Within radiation oncology, we hypothesize treatment planning to be a clinical task that necessitates System 2 reasoning because of the complex nature of optimization tradeoffs. 

We operationalized six categories of System 2 cognitive processing derived from the dual-process literature: problem decomposition (breaking complex objectives into sub-goals), prospective verification (checking constraints before action), self-correction (revising approach after recognizing errors), mathematical reasoning (explicit numerical computation), trade-off deliberation (weighing competing objectives), and forward simulation (predicting dosimetric consequences of proposed actions). This categorization was adapted from previous work in System 2 LLM classification \cite{r50}. We additionally quantified format errors, defined as malformed structured output that failed to parse. 

We employed a hybrid automated-manual approach for content analysis. A custom script performed initial detection of System 2 cognitive processes using keyword and phrase pattern matching. The script searched for linguistic markers associated with each cognitive category (Table 1). A random sample of 10\% of automated classifications was manually verified assess concordance. Format errors were defined as malformed structured output that failed JSON parsing.

\begin{table}[h]
\caption{Cognitive categories identified in the reasoning analysis, including linguistic markers and example terminology.}
\label{tab:cognitive_markers}
    \renewcommand{\arraystretch}{1.4} 
    
    \begin{tabular}{p{3.5cm} p{4.5cm} p{5cm}}
        \toprule
        \textbf{Cognitive category} & \textbf{Linguistic marker} & \textbf{Example words} \\
        \midrule
        
        Problem decomposition & 
        Planning language & 
        `First', `then', `next', `I will start by' \\
        
        Prospective verification & 
        Conditional statements with constraint references & 
        `If... then', `would exceed', `checking whether', `to make sure V12Gy stays under' \\
        
        Self-correction & 
        Revision language & 
        `reverting', `instead', `previous attempt', `I will revise', `This assumption was incorrect' \\
        
        Mathematical reasoning & 
        Numerical expressions and calculations & 
        `delta', `fraction', `greater than', `increase from X to Y' \\
        
        Trade-off deliberation & 
        Comparative language involving competing objectives & 
        `balance', `prioritize', `versus', `at the cost of' \\
        
        Forward simulation & 
        Predictive language & 
        `will cause', `expected to', `will result in' \\
        
        \bottomrule
    \end{tabular}
\end{table}

\subsection{Statistical Analysis and Experimental Design}\label{subsec5}
We used paired, non-parametric Wilcoxon signed-rank tests for all plan-to-plan comparisons, with statistical significance defined as p \textless\ 0.05. The choice of a non-parametric approach was supported by Shapiro-Wilk testing of paired differences and by visual inspection of Q-Q plots. Multiplicity correction was performed using the Benjamini-Hochberg (BH) procedure to control the false discovery rate (FDR) at q \textless\ 0.05 within two pre-specified hypothesis families: primary endpoints (target coverage, maximum dose, conformity index, gradient index) and secondary endpoints (all seven OARs).

Data were visualized using violin plots for each dosimetric endpoint. Individual patient values were displayed as jittered points, with interquartile range (IQR) boxes and median values overlaid. Significance brackets were applied only to comparisons that remained significant after BH correction. All statistical analyses and visualizations were performed in R using the tidyverse, ggsignif, and ggbeeswarm packages.

\section{Results}\label{sec3}
\subsection{Target coverage and dose homogeneity}\label{subsec1}
Both model variants met clinical acceptance criteria for target dosimetry (Figure 2). The reasoning model achieved median PTV coverage of 96.8\% (IQR: 95.9-97.4\%), compared to 96.5\% (IQR: 95.6-97.2\%) for clinical plans. The non-reasoning model achieved median coverage of 96.2\% (IQR: 95.1-97.0\%). All three groups maintained median maximum doses below the 21.6 Gy threshold, with the reasoning model producing a distribution most closely aligned with clinical plans. No patient in either AI cohort fell below the 95\% coverage threshold.

\begin{figure}[H]
  \centering
  \includegraphics[width=0.9\textwidth]{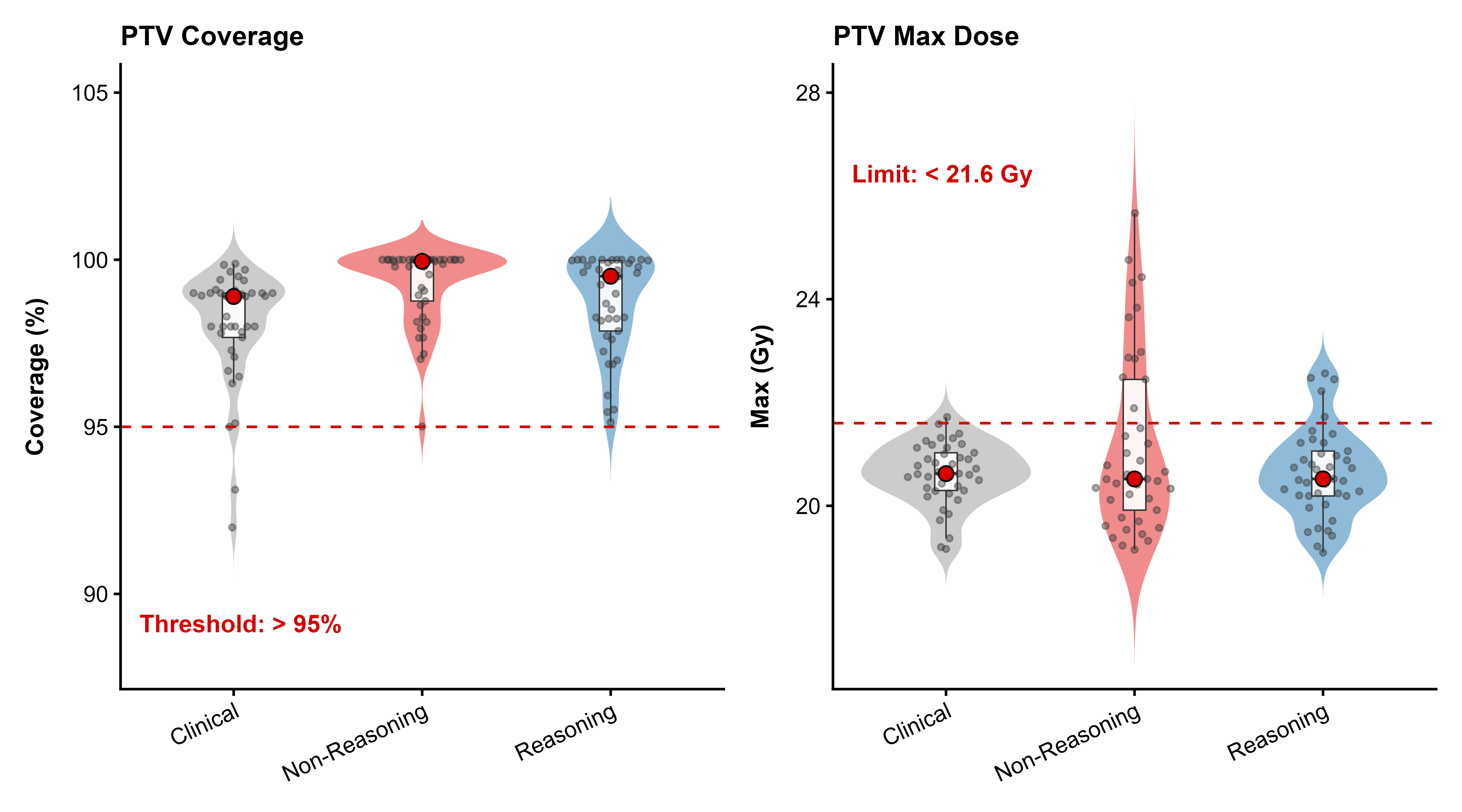}
  \caption{PTV coverage (left) and maximum dose (right) for clinical plans (grey), non-reasoning model (red), and reasoning model (blue). Violin contours represent kernel density estimates. White boxes indicate IQR; red points indicate median values; black points represent individual patients (n = 41). Dashed lines denote clinical acceptance thresholds (coverage > 95\%; maximum dose \textless\ 21.6 Gy). Brackets indicate comparisons between reasoning and clinical groups that remained significant after BH correction (q \textless\ 0.05).}
  \label{fig:2}
\end{figure}

\subsection{OAR Sparing}\label{subsec2}
For the brainstem and optic chiasm, the reasoning model produced dose distributions comparable to clinical plans, with all cases remaining below institutional tolerance thresholds. Normal brain exposure, quantified as V12Gy (volume receiving at least 12 Gy), remained within the clinical limit of 10 cc for most cases across all three cohorts. No significant differences were observed between the reasoning model and clinical plans for any central OAR endpoint after Benjamini-Hochberg correction.

Lateral OARs, including the bilateral optic nerves and cochleae, were assessed against a maximum dose threshold of 9 Gy (Figure 4). Both AI variants maintained doses below this limit across all structures. The reasoning model achieved significantly lower doses to the right cochlea compared to clinical plans (p = 0.022 after BH correction). Doses to the left cochlea, right optic nerve, and left optic nerve did not differ significantly between the reasoning model and clinical plans.

\begin{figure}[H]
  \centering
  \includegraphics[width=0.9\textwidth]{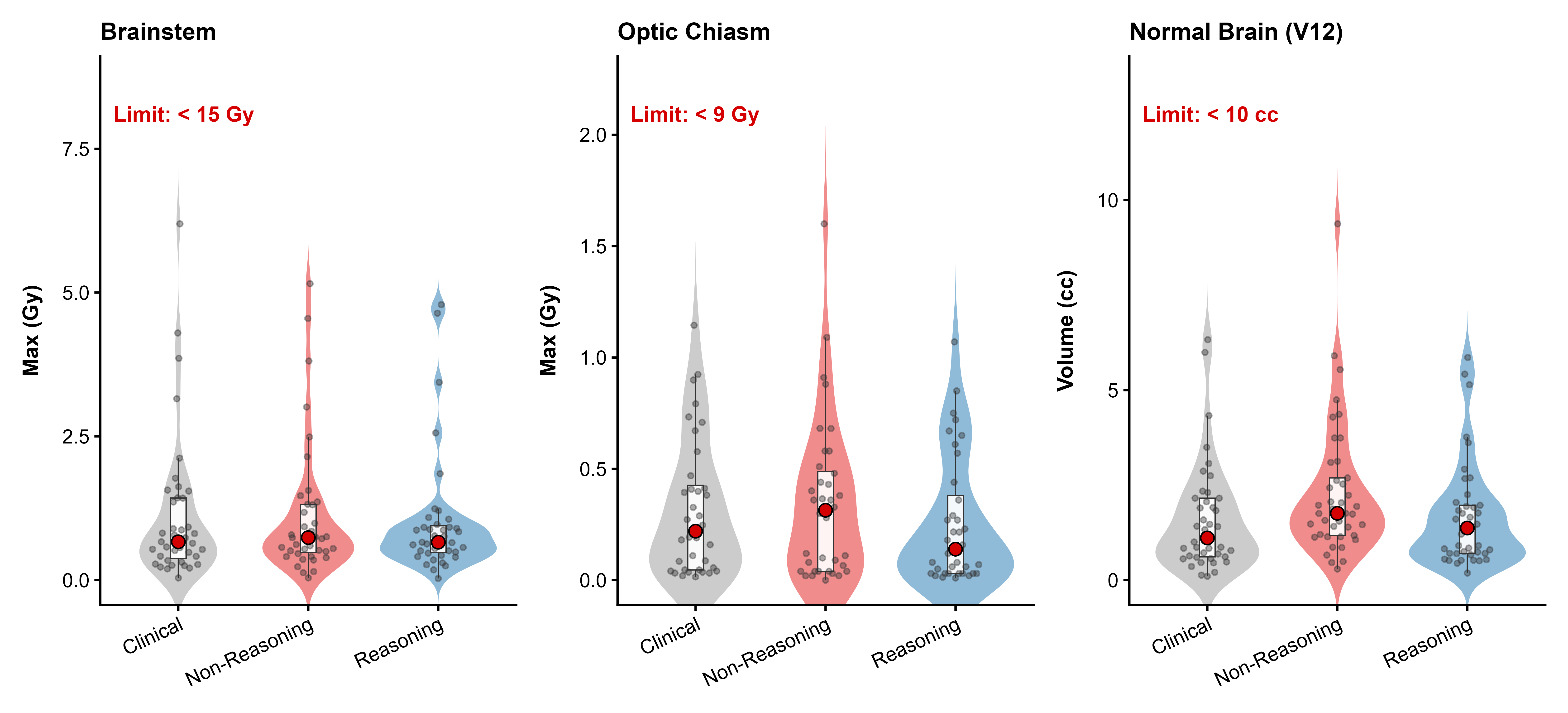}
  \caption{Maximum doses to brainstem and optic chiasm, and V12Gy for normal brain (defined as brain minus gross tumor volume (GTV)) across clinical plans (grey), non-reasoning model (red), and reasoning model (blue). Violin contours represent kernel density estimates. White boxes indicate IQR; red points indicate median values; black points represent individual patients (n = 41). Brackets indicate comparisons between reasoning and clinical groups that remained significant after BH correction (q \textless\ 0.05).}
  \label{fig:3}
\end{figure}

\begin{figure}[H]
  \centering
  \includegraphics[width=0.9\textwidth]{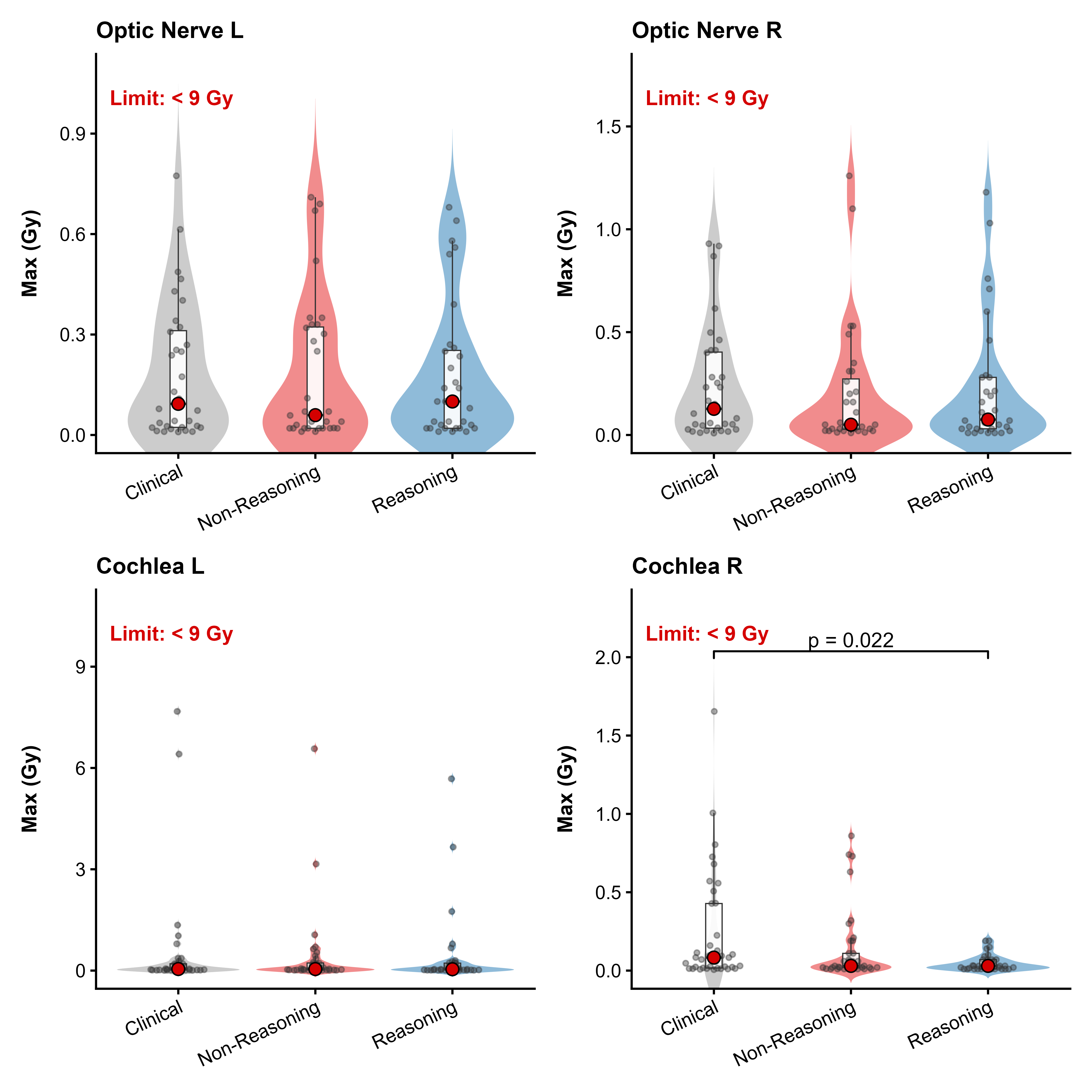}
  \caption{Maximum doses to bilateral optic nerves and cochleae across clinical plans (grey), non-reasoning model (red), and reasoning model (blue). Violin contours represent kernel density estimates. White boxes indicate IQR; red points indicate median values; black points represent individual patients (n = 41). The reasoning model achieved significantly lower right cochlear doses compared to clinical plans. All other comparisons were not significant.}
  \label{fig:4}
\end{figure}

\subsection{Response to human-in-the-loop refinement}\label{subsec3}
We first assessed whether the AI agent could respond appropriately to human feedback regarding plan conformity. Both model variants demonstrated statistically significant improvement in conformity index (CI) following a natural language refinement prompt by human (Figure 5). Both models achieved statistically significant improvements in CI (reasoning: p \textless\ 0.001, non-reasoning: p = 0.007). The smaller p-value for the reasoning model reflects greater consistency of improvement across patients (lower variance), as evidenced by narrower confidence intervals, rather than necessarily greater magnitude of improvement.

After refinement, the reasoning model achieved a median CI that closely approximated the clinical benchmark, whereas the non-reasoning model, although significantly improved, remained further from clinical values. Both models interpreted and acted on natural language feedback, but the reasoning model achieved the larger conformity gain.

\begin{figure}[H]
  \centering
  \includegraphics[width=0.9\textwidth]{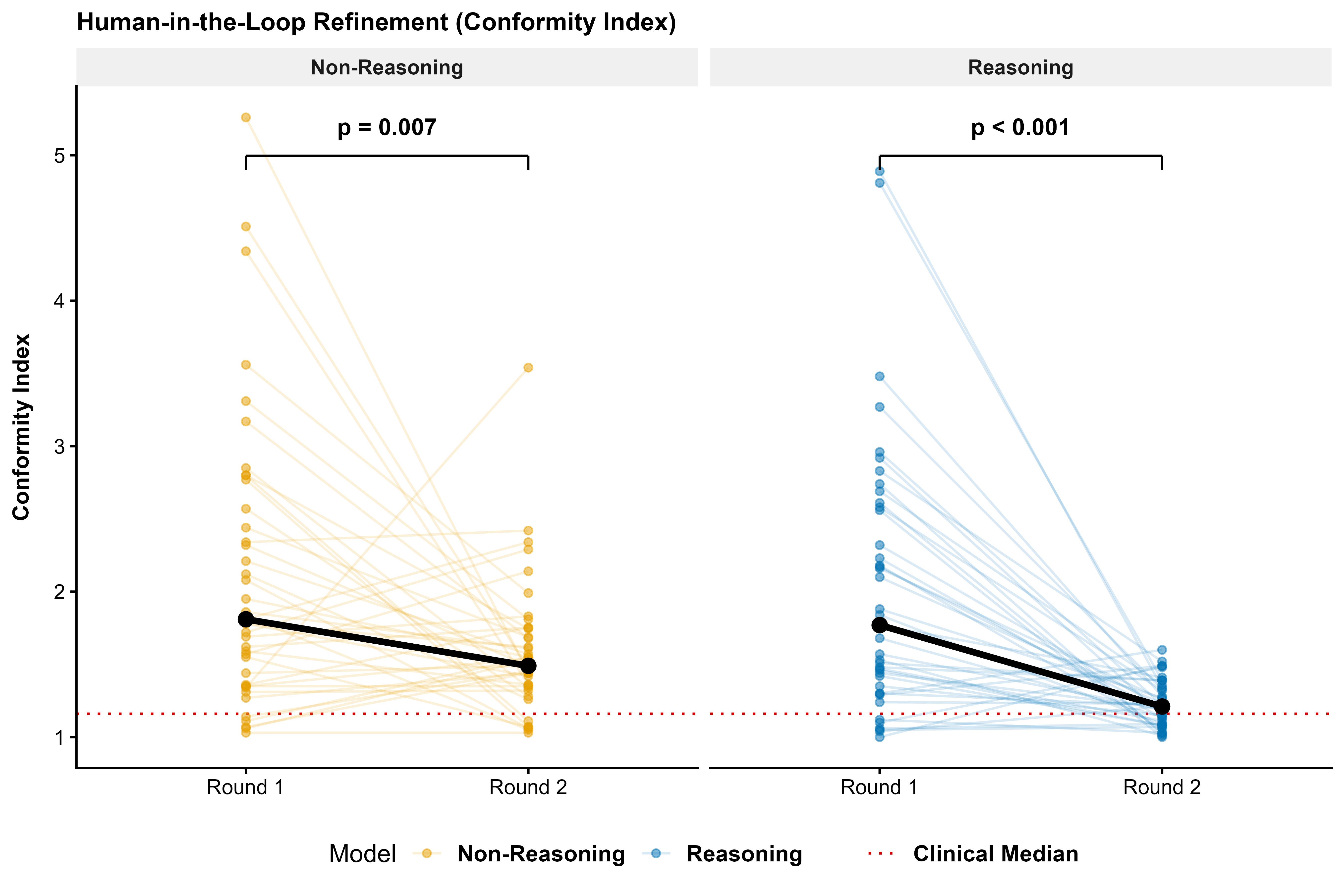}
  \caption{Conformity index values before (Round 1) and after (Round 2) the refinement prompt for reasoning (blue) and non-reasoning (yellow) model variants. Both models showed significant improvement following refinement (non-reasoning: p = 0.007; reasoning: p \textless\ 0.001, paired Wilcoxon signed-rank test). The reasoning model achieved median conformity approaching clinical benchmark values. Boxes indicate IQR; horizontal lines indicate median; whiskers extend to a 1.5 multiple of IQR.}
  \label{fig:5}
\end{figure}

\subsection{Paired comparison of reasoning agent versus clinical plans}\label{subsec4}
We performed paired difference analysis comparing the reasoning agent to clinical plans across all dosimetric endpoints (Figure 6). The reasoning agent demonstrated dosimetric outcomes comparable to clinical plans across primary target coverage metrics, with no statistically significant differences detected. PTV coverage did not differ significantly (p = 0.21), nor did maximum dose (p = 0.98). Plan quality metrics, including conformity index (p = 0.23) and gradient index (p = 0.71), did not differ significantly between the reasoning agent and human planners. For OAR endpoints, the reasoning agent achieved comparable or superior performance across all structures. The right cochlea was the only structure demonstrating a statistically significant difference, favoring SAGE (p = 0.022 after BH correction). 

\begin{figure}[H]
  \centering
  \includegraphics[width=0.9\textwidth]{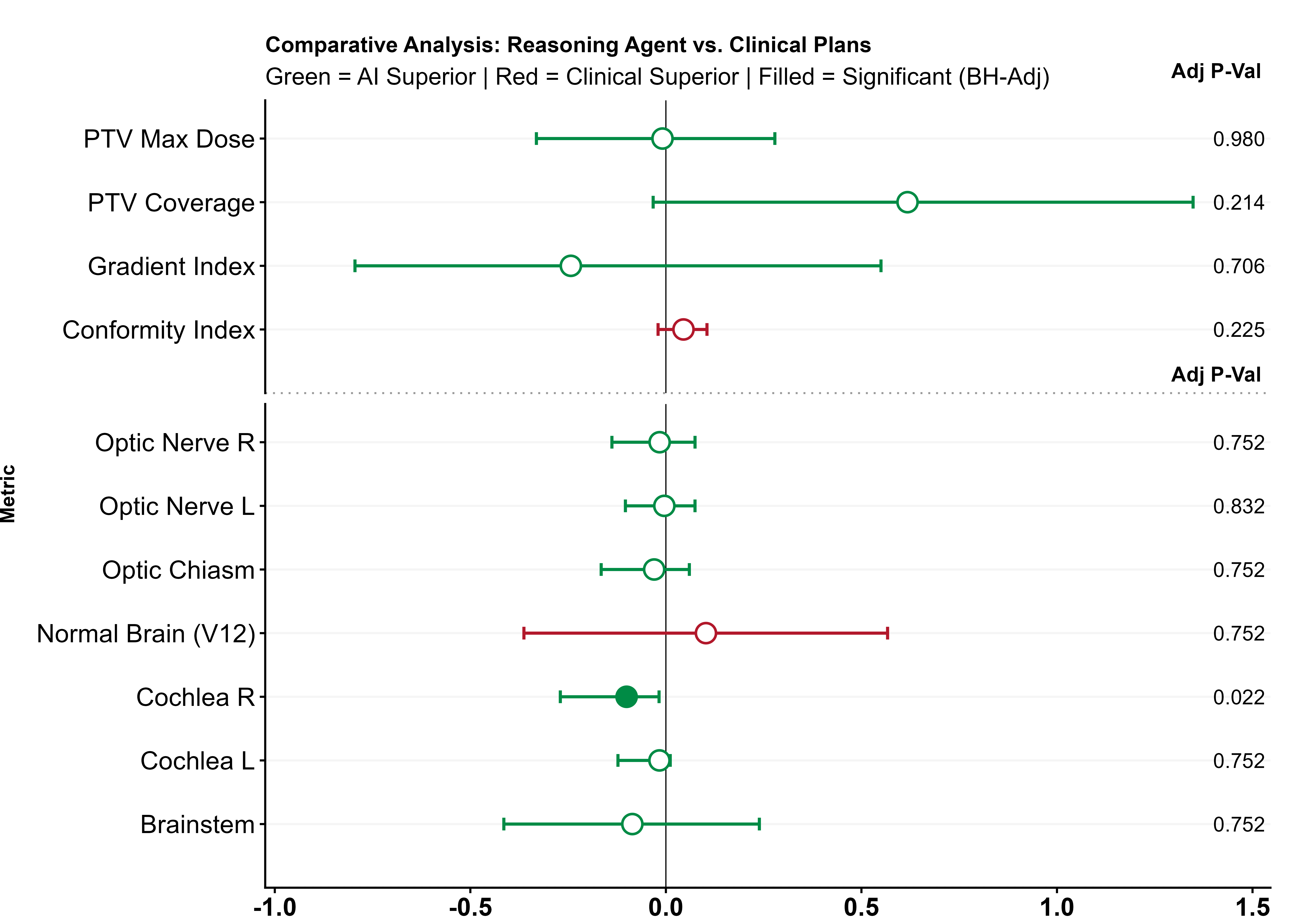}
  \caption{Median paired differences (reasoning minus clinical) across all dosimetric endpoints. For OAR doses and plan quality metrics (CI, GI), values left of zero indicate AI superiority (lower doses, better conformity/gradient); for PTV coverage, values right of zero indicate AI superiority (higher coverage). n = 41 patients; error bars represent 95\% confidence intervals; paired Wilcoxon signed-rank tests with BH correction for secondary endpoints. The reasoning agent demonstrated significantly improved right cochlear sparing (q \textless\ 0.05). No significant differences were observed for target coverage, maximum dose, conformity index, gradient index, or other OAR endpoints.}
  \label{fig:6}
\end{figure}

\subsection{Mechanistic differences between reasoning and non-reasoning agents}\label{subsec5}
Problem decomposition, prospective verification, and self-correction were detected exclusively in the reasoning model (n = 537, 457, and 735 instances, respectively). The non-reasoning model produced zero instances of these behaviors across all 41 patients and all optimization iterations. The remaining cognitive processes, while present in both models, showed similar asymmetric distributions. Mathematical reasoning occurred primarily in the reasoning model (1,162 vs. 49 instances; 96\% share). Trade-off deliberation (609 vs. 7 instances; 99\% share) and forward simulation (1,888 vs. 58 instances; 97\% share) followed the same pattern. The reasoning model dominated across all six cognitive categories. These behavioral differences translated to measurable reliability improvements. The reasoning model produced five-fold fewer format errors than the non-reasoning model (25 vs. 122 total; median 0 vs. 3 per patient).

\begin{figure}[H]
  \centering
  \includegraphics[width=0.9\textwidth]{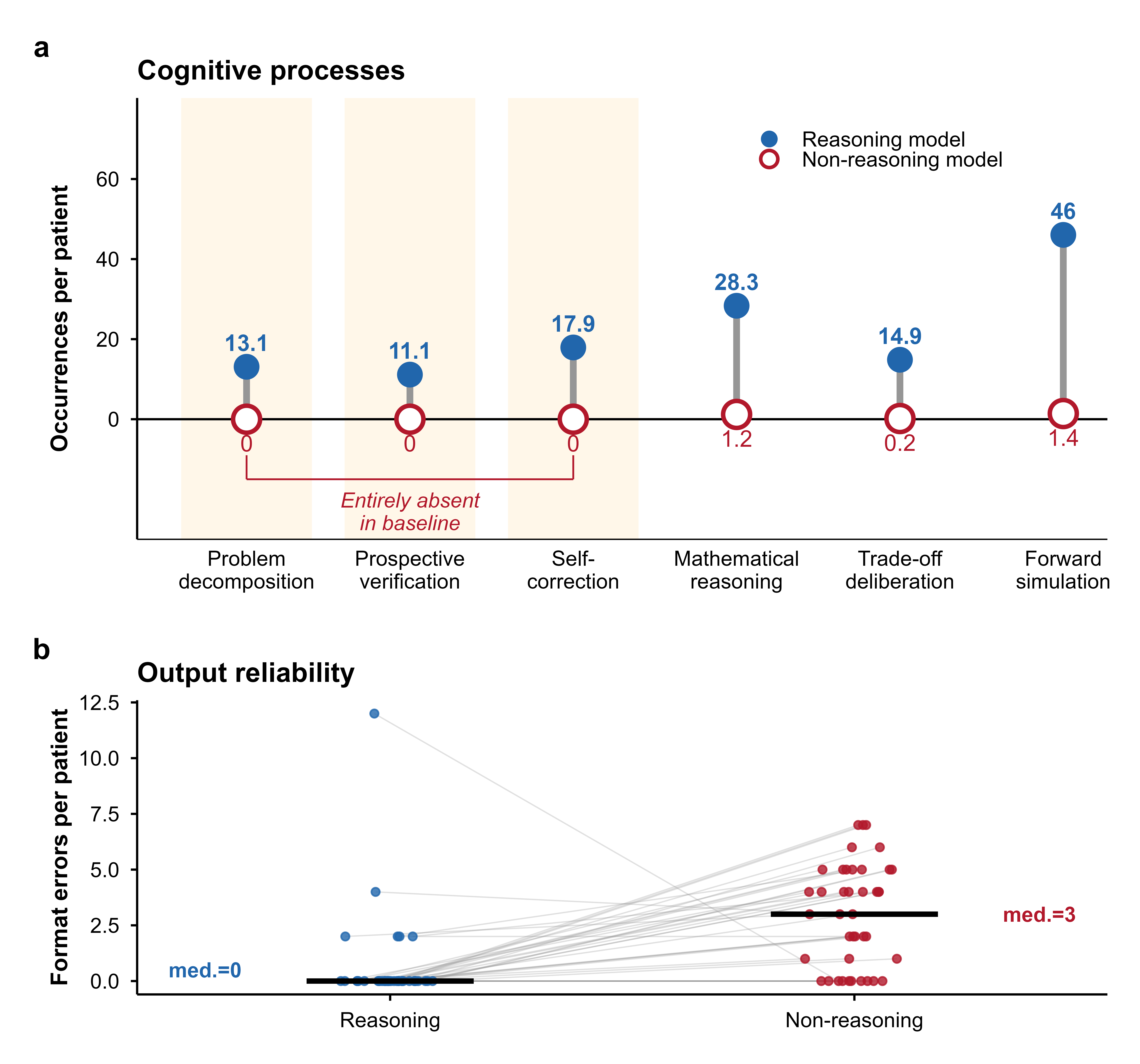}
  \caption{System 2 cognitive processes (problem decomposition, prospective verification, self-correction, mathematical reasoning, trade-off deliberation, forward simulation) were quantified across all patients. Three processes were detected exclusively in the reasoning model; the remaining three occurred in both models but were concentrated in the reasoning agent. Output reliability was assessed by format errors per patient; the reasoning model achieved median 0 errors versus median 3 for non-reasoning.}
  \label{fig:7}
\end{figure}

\section{Discussion}\label{sec4}

This study demonstrates that an AI planning agent equipped with System 2 reasoning capabilities can produce SRS plans that are equivalent to those generated by experienced human dosimetrists. In the case of cochlear sparing, the reasoning agent achieved statistically superior performance. The cognitive architecture of an AI system may be as consequential for clinical performance as the quality of its training data or the sophistication of its optimization algorithms.

Our comparison of System 1 and System 2 thinking within the same agent framework distinguishes this work from prior autoplanning studies. Both SAGE variants met clinical constraints for target coverage and OAR dose limits. However, the reasoning model achieved significantly lower doses to the right cochlea (p = 0.022 after BH FDR correction), despite both AI and human plans remaining well below the 9 Gy tolerance threshold. 

This difference may potentially reflect distinct optimization strategies. Human planners operate under pragmatic constraints; once an OAR meets its dose threshold, clinical workflow pressures discourage further optimization \cite{r5}. Shift schedules, competing responsibilities, and cognitive fatigue all favor satisficing behavior. The reasoning agent, free from these pressures, appears to have implemented the ALARA (as low as reasonably achievable) principle more completely. The cochlear result does not indicate a failure of human planning. Rather, it may reflect an inherent limitation of the satisficing heuristic in dose optimization. Whether this additional OAR sparing translates to reduced late toxicity remains to be determined in prospective studies.

The CI results merit particular consideration \cite{r51}. Given that SRS delivers high doses in a single fraction to targets surrounded by sensitive structures, dose conformity is of primary clinical importance. Our human-in-the-loop experiment demonstrated that both AI variants improved their conformity indices following a natural language refinement prompt. The reasoning model exhibited more pronounced improvement compared to the non-reasoning model, with median values approaching clinical benchmarks. Spatial reasoning in three dimensions, specifically the manipulation of dose distributions around geometrically complex targets, appears to benefit from deliberative processing. The reasoning agent's use of spherical rind structures to improve conformity exemplifies the strategic planning approaches employed by experienced dosimetrists, approaches that simpler models appear unable to generalize. This agrees with several previous works in other domains that have demonstrated the improved performance of reasoning or System 2 models at three dimensional tasks \cite{r52,r53,r54}.

The deployment of autonomous reasoning tools such as SAGE suggest a potential reorganization of roles in future radiation oncology treatment planning. If optimization can be reliably delegated to a System 2 agent, dosimetrists and physicists may focus on clinical judgment, quality assurance, and strategic decisions that remain beyond current AI capabilities. The human-in-the-loop architecture represents a model for human-AI collaboration in which each contributes according to its strengths. Humans provide clinical context, identifies when conformity requires refinement, and exercises final approval. 

Several limitations should be acknowledged. First, our comparisons were based on non-inferiority rather than formal equivalence testing with pre-specified margins. The observed associations between reasoning behaviors and dosimetric outcomes do not establish causation; differences could reflect model architecture, training data, or prompting rather than intrinsic reasoning capabilities. Our 41-patient cohort derives from a single institution. While this sample provides adequate statistical power for the comparisons we undertook, external validation across centers with different planning conventions and beam configurations will be necessary before broader claims of generalizability can be supported. The computational demands of System 2 reasoning, particularly the inference-time costs of models such as QwQ-32B, remain substantial and may limit deployment in resource-constrained settings. The reasoning model required approximately threefold longer inference time per plan compared to the non-reasoning variant. Although we demonstrated non-inferiority across a comprehensive set of dosimetric endpoints, the ultimate outcome of interest, patient survival and toxicity, lies beyond the scope of this retrospective analysis. Finally, the asymmetric cochlear finding warrants comment. We cannot definitively explain this laterality. Possible contributors include the spatial distribution of tumor locations in our cohort, which may have placed more lesions in proximity to the right cochlea; systematic asymmetries in beam geometry inherited from the clinical plans; or statistical variation inherent to multiple comparisons across a 41-patient sample.

This study is best understood as a behavioral analysis of AI architecture rather than a clinical efficacy trial. The effect sizes of behavioral differences suggest that the distinction between System 1 and System 2 architectures is robust regardless of cohort scale. We cannot fully disentangle the contribution of explicit reasoning from other differences between the two LLMs, including training data composition and model architecture. An important practical consideration is that the reasoning model is not universally superior for all treatment planning scenarios. Our findings suggest the reasoning model's advantages emerge primarily in cases involving subtle dosimetric trade-offs or when transparency in the optimization process is clinically valuable, such as understanding why certain OAR doses could not be further reduced. For routine cases with well-defined constraints and limited degrees of freedom, the non-reasoning model often achieves clinically acceptable plans more efficiently. Inference times were approximately threefold faster with reduced computational cost. The choice between models should therefore be guided by case complexity, the need for optimization transparency, and available computational resources.
Future work will include prospective, multi-institutional validation with outcome assessment; reader studies comparing oncologist and physicist ratings of AI-generated versus human plans; and extension of the reasoning framework to other complex indications including spine SRS, multiple brain metastases, and extracranial stereotactic body radiotherapy (SBRT).

\section{Conclusion}\label{sec5}
We have demonstrated that a System 2 reasoning agent can generate stereotactic radiosurgery plans that meet or exceed the quality of those produced by experienced human planners. Across 41 patients, the reasoning variant of SAGE achieved equivalent performance on all primary dosimetric endpoints, including target coverage, maximum dose, conformity index, and gradient index. On one secondary endpoint, right cochlear dose, the reasoning agent achieved statistically superior sparing compared to clinical plans. These results were accompanied by qualitatively distinct planning behavior: the reasoning model exhibited constraint verification, causal explanation, and iterative memory reference patterns consistent with deliberative cognition, whereas the non-reasoning model exhibited reactive parameter adjustment without explicit justification. The distinction between reasoning and non-reasoning architectures has practical consequences for both plan quality and interpretability. As reasoning models improve in capability and efficiency, their integration into radiation oncology workflows should be actively considered.

\bibliography{bibliography}

\end{document}